\documentclass[conference]{IEEEtran}
\IEEEoverridecommandlockouts
\usepackage{cite}
\usepackage{amsmath,amssymb,amsfonts}
\usepackage{algorithm}
\usepackage{algorithmic}
\usepackage{graphicx}
\usepackage{textcomp}
\usepackage{xcolor}
\usepackage{booktabs} 
\usepackage{multirow}
\usepackage{url}
\usepackage{colortbl}
\usepackage[T1]{fontenc}
\usepackage[utf8]{inputenc}
\usepackage{threeparttable}
\usepackage{tikz}
\usetikzlibrary{positioning, arrows.meta}
\usepackage{enumitem}

\begin{document}

\title{VARS-FL: Validation-Aligned Client Selection for Non-IID Federated Learning in IoT Systems}

\author{\IEEEauthorblockN{Mohamed Lakas\IEEEauthorrefmark{1},
Mohamed Amine Ferrag\IEEEauthorrefmark{1}\IEEEauthorrefmark{2}}
\IEEEauthorblockA{\IEEEauthorrefmark{1}College of Information Technology, United Arab Emirates University, UAE}
\IEEEauthorblockA{\IEEEauthorrefmark{2}
Corresponding author: \texttt{mohamed.ferrag@uaeu.ac.ae}
}

}
\maketitle


\begin{abstract}
Federated learning (FL) systems typically employ stateless client selection, treating each communication round independently and ignoring accumulated evidence of client contribution quality. Under non-IID data, this leads to slow convergence and unstable training, particularly when selection relies on local proxies (e.g., training loss) that are misaligned with the global optimization objective. These challenges are especially pronounced in Internet of Things (IoT) and Industrial IoT (IIoT) environments, where data is highly heterogeneous and distributed across devices observing different traffic patterns. In this paper, we propose VARS-FL (Validation-Aligned Reputation Scoring for Federated Learning), a client selection framework that quantifies each client’s contribution using the reduction in server-side validation loss induced by its update. These per-round signals are aggregated into a Reputation score that combines a sliding-window average of recent contributions with a logarithmically scaled participation term, enabling robust exploration–exploitation selection. VARS-FL requires no changes to local training or aggregation and remains fully compatible with standard FedAvg. We evaluate VARS-FL on a 15-class non-IID IoT intrusion detection task using the Edge-IIoTset dataset with 100 clients across multiple seeds, comparing it against FedAvg, Oort, and Power-of-Choice. VARS-FL consistently improves accuracy, F1-Macro, and loss, while accelerating convergence (up to 36\% fewer rounds to reach 80\% accuracy). These results demonstrate that validation-aligned, history-aware client selection provides a more reliable and efficient training process for federated learning in heterogeneous IoT environments.
\end{abstract}



\section{Introduction}

Federated learning (FL) enables multiple clients to collaboratively train a shared model without exchanging raw data, making it particularly well-suited for privacy-sensitive environments such as mobile systems and healthcare applications~\cite{mcmahan2017communication,kairouz2021advances}. This paradigm has been successfully deployed in data-distributed scenarios, including mobile keyboard next-word prediction and other forms of on-device personalization, and has gained increasing attention in domains where data sharing is restricted by privacy regulations and policy constraints, such as healthcare and cybersecurity~\cite{ferrag2022edge}.

The Internet of Things (IoT) and Industrial Internet of Things (IIoT) represent especially compelling application domains for FL. These environments comprise large numbers of geographically distributed devices that generate highly heterogeneous, localized data streams. Each node observes only a limited, potentially biased subset of the global data distribution — e.g., specific traffic patterns or attack types — yet, collectively, these nodes contain the information necessary to train effective intrusion detection systems. However, centralizing such data is often impractical due to bandwidth limitations, data sovereignty requirements, and the sensitive nature of industrial network traffic~\cite{chung2026decentralized}. In this context, FL provides a principled solution by enabling devices to perform local training and share model updates rather than raw data, thereby enabling collaborative learning while preserving data locality and privacy~\cite{hamouda2023ppss,hamouda2024revolutionizing}.
\textcolor{black}{
A major drawback of FL is the significant communication overhead it introduces. In each round, participating clients upload gradients (or updated model parameters) which can be prohibitively large, particularly for bandwidth-limited clients and deep models \cite{konevcny2016federated}. In IIoT environments, edge devices typically operate over constrained wireless links (e.g., LPWAN, narrowband LTE, or shared Wi-Fi), which makes communication overhead a key bottleneck. To mitigate this cost, several complementary approaches have been proposed. First, aggregation and system mechanisms to achieve robust, privacy-preserving updates at scale, e.g., via secure aggregation \cite{bonawitz2017practical}. Second, update compression to reduce message sizes through quantization and sparsification techniques \cite{konevcny2016federated, alistarh2017qsgd, aji2017sparse}. Finally, client selection to reduce the number of clients participating per round by scheduling a subset expected to contribute most under resource and data heterogeneity \cite{nishio2019client, shi2021joint}.}

FedAvg and most standard FL algorithms adopt \emph{stateless} client selection, in which each communication round is treated independently, and selection decisions do not account for historical client behavior. As a result, future participation is not informed by a client's past contribution to global model improvement \cite{quan2025federated,pfeiffer2023federated}. In realistic deployments, client data exhibit significant heterogeneity in quality, quantity, and class distribution, and without tracking past contributions, the FL server cannot distinguish between clients that consistently provide beneficial updates and those that are noisy or uninformative \cite{uddin2025systematic,fu2023client}. This limitation is particularly critical in IIoT intrusion detection settings, where the distribution of attack types varies substantially across nodes; for example, a sensor monitoring industrial SCADA traffic may predominantly observe Ransomware and Backdoor attacks, whereas a gateway node may encounter mostly DDoS traffic. Under such non-IID conditions, clients holding rare but important classes are often under-selected despite being essential for improving global model coverage \cite{thakur2025green}. This issue is further exacerbated under partial participation: in a federation with 100 clients and a 10\% selection rate per round, each client has only a 10\% probability of being selected in any given round, regardless of its utility, leading to inefficient training and suboptimal generalization \cite{benarba2025bias}.

A further, often underappreciated challenge in federated learning is \emph{objective misalignment}: when client selection is driven by local proxies (e.g., local training loss), the server may over-select clients that optimize their own local objectives while contributing little, or even negatively, to the global objective \cite{ye2023heterogeneous}. This misalignment arises because local loss reflects the client-specific data distribution rather than the global data distribution. Under non-IID conditions, a client holding rare or highly skewed classes may exhibit high local loss not because it is globally informative, but because the current global model is poorly adapted to its distribution. Prioritizing such clients can bias the global model toward their local data, degrading overall generalization \cite{almanifi2023communication}. Moreover, this effect accumulates over rounds: repeated selection based on local loss progressively steers the global model into suboptimal directions, thereby degrading the initialization for subsequent updates. Empirically, this phenomenon is evident in methods such as Oort~\cite{lai2021oort}, which favor high local-loss clients and consequently exhibit unstable training dynamics, including persistent accuracy oscillations and failure to reach the 80\% accuracy threshold within 100 rounds across multiple random seeds \cite{beltran2023decentralized}.

In this paper, we present \textit{VARS-FL}, a client selection framework that simultaneously addresses both the statelessness and objective-misalignment challenges in federated learning, with a particular focus on heterogeneous Internet of Things (IoT) environments, without modifying local training, aggregation, or communication protocols. Unlike prior approaches that rely on local proxies, VARS-FL directly measures each client's contribution using server-side validation loss, providing a signal that is inherently aligned with the global optimization objective. This validation-driven signal identifies clients that contribute to global generalization, requires no additional client-side disclosures beyond model updates, and is accumulated over time into a per-client \emph{Reputation} score that captures consistent contribution across rounds.

The main contributions of this paper are as follows:
\begin{enumerate}[label=(\roman*)]
    \item We propose a \emph{validation-loss-based quality score} that directly quantifies each client's per-round contribution to global model improvement using a server-side validation set, ensuring alignment with the global objective by construction, particularly under non-IID data distributions common in IoT systems;

    \item We introduce a \emph{Reputation scoring mechanism} that combines a sliding-window average of recent quality scores with a logarithmically scaled participation term, enabling selection decisions to prioritize consistent contribution rather than mere participation frequency in dynamic and heterogeneous IoT environments;

    \item We develop a \emph{multi-armed bandit (MAB)-based selection strategy} that formulates client selection as an explore--exploit problem, balancing the discovery of informative clients with the repeated selection of those that have demonstrated reliable global benefit in non-stationary IoT settings;

    \item We conduct an extensive evaluation of VARS-FL on the Edge-IIoTset Cyber Security Dataset~\cite{ferrag2022edge} in a 100-client, non-IID IoT intrusion detection setting across multiple random seeds.
\end{enumerate}

The remainder of this paper is organized as follows. Section~\ref{sec:related} reviews related work on federated learning under data heterogeneity, client selection strategies, reputation and trust mechanisms, bandit-based optimization, and FL-based intrusion detection. Section~\ref{sec:system} presents the VARS-FL framework, including data preprocessing, model architecture, validation-loss-based quality scoring, reputation aggregation, the explore--exploit selection policy, and complexity analysis. Section~\ref{sec:setup} describes the experimental setup, including the dataset, non-IID partitioning scheme, baselines, and evaluation metrics. Section~\ref{sec:results} reports and analyzes the experimental results in terms of overall performance, convergence speed, training dynamics, and per-class recall. Finally, Section~\ref{sec:conclusion} concludes the paper and outlines directions for future work.

\begin{table*}[t]
\centering
\caption{Comparison of client selection methods across key design dimensions}
\label{tab:related}
\begin{threeparttable}
\setlength{\tabcolsep}{4pt}
\renewcommand{\arraystretch}{1.2}
\begin{tabular}{lcccccc}
\toprule
\textbf{Method} 
& \textbf{Utility Signal} 
& \textbf{History Aware} 
& \textbf{Global Alignment} 
& \textbf{Non-IID Robustness} 
& \textbf{Minority Support} 
& \textbf{IoT/Sec. Eval.} \\
\midrule
FedAvg~\cite{mcmahan2017communication}
  & None        & No  & --        & No      & No & No \\
FedProx~\cite{li2020fedprox}
  & None        & No  & --        & Partial & No & No \\
PoC~\cite{cho2022power}
  & Local loss  & No  & Indirect  & Partial & No & No \\
Oort~\cite{lai2021oort}
  & Local loss  & No  & Indirect  & No      & No & No \\
Active FL~\cite{goetz2019active}
  & Grad.\ norm & No  & Partial   & No      & No & No \\
FedCor~\cite{tang2022fedcor}
  & Grad.\ corr.& No  & Partial   & Partial & No & No \\
FLTrust~\cite{cao2021fltrust}
  & Cosine sim. & No  & Partial   & Partial & No & No \\
\midrule
\textbf{VARS-FL (Ours)}
  & Val.\ loss  & Yes & \textbf{Explicit} & \textbf{Yes} & \textbf{Yes} & \textbf{Yes} \\
\bottomrule
\end{tabular}

\textit{Utility Signal}: metric used to estimate client contribution (e.g., validation loss, gradient norm). 
\textit{History Aware}: whether past client performance is incorporated. 
\textit{Global Alignment}: whether the signal directly reflects improvement in the global objective. 
\textit{Non-IID Robustness}: reported stability under heterogeneous data distributions. 
\textit{Minority Support}: explicit consideration of rare or underrepresented classes. 
\textit{IoT/Sec.\ Eval.}: evaluation on IoT or cybersecurity datasets.

\end{threeparttable}
\end{table*}

\section{Related Work}
\label{sec:related}

This section reviews prior work relevant to VARS-FL across five complementary directions. We first discuss approaches that address data heterogeneity in federated learning, focusing on optimization-based methods that mitigate client drift. Then, we examine existing client selection strategies and highlight their reliance on local proxy signals. Next, we review reputation and trust mechanisms, followed by bandit-based formulations for adaptive client selection. Finally, we cover applications of federated learning in IoT security and intrusion detection. Throughout, we emphasize the gap between local utility estimation and global objective alignment, which motivates the design of VARS-FL.

\subsection{Federated Learning Under Data Heterogeneity}

Data heterogeneity is a fundamental challenge in federated learning. In realistic deployments, client data distributions differ in label composition, feature space, and sample volume, a setting commonly referred to as non-IID (non-independent and identically distributed). Li~et~al.~\cite{li2020convergence} show that such heterogeneity induces \emph{client drift}, where local updates move the global model toward client-specific optima, thereby degrading global generalization and slowing convergence even under standard FedAvg~\cite{mcmahan2017communication}.

Several approaches mitigate client drift by modifying the local optimization objective. FedProx~\cite{li2020fedprox} introduces a proximal term to limit deviation from the global model. SCAFFOLD~\cite{karimireddy2020scaffold} employs control variates to correct gradient drift, while MOON~\cite{li2021model} leverages contrastive learning to align local and global representations. Although these methods reduce the adverse effects of heterogeneous updates, they do not address \emph{client selection}; instead, they assume uniform participation and focus on stabilizing the optimization process. VARS-FL is complementary to these approaches. Rather than mitigating the impact of arbitrary client selection, VARS-FL reduces the likelihood of selecting clients whose updates are misaligned with the global objective. As such, it operates at the selection stage, orthogonal to optimization-based methods, and can be integrated with techniques such as FedProx or SCAFFOLD without modification.

\subsection{Client Selection Strategies}

Client selection plays a critical role in federated learning, particularly under heterogeneous data distributions. Standard FedAvg~\cite{mcmahan2017communication} selects clients uniformly at random, ignoring both update quality and data distribution, which leads to statistical inefficiency. For instance, in a federation with 100 clients and a 10\% participation rate, a client holding rare but important data has only a 10\% probability of being selected in any round, regardless of its utility. Prior work propose various strategies to estimate client utility, typically using local proxies. Power-of-Choice~\cite{cho2022power} prioritizes clients with high local loss, assuming they contain under-learned data. However, this conflates local difficulty with global utility, as high local loss may arise from distributional divergence rather than informative content. Active federated learning~\cite{goetz2019active} uses gradient norms as a proxy for informativeness, while FedCor~\cite{tang2022fedcor} models inter-client loss correlations via Gaussian processes to guide selection.

Oort~\cite{lai2021oort} combines statistical utility and system efficiency, using local loss as the utility signal, while FedScale~\cite{lai2022fedscale} extends this paradigm to large-scale system-aware benchmarking. FedCS~\cite{nishio2019client} focuses on resource-constrained environments, selecting clients based on latency constraints~\cite{shi2021joint}.

A fundamental limitation of these methods is their reliance on \emph{local proxies}---such as local loss, gradient norm, or correlation---as surrogates for global utility. These signals reflect properties of local data distributions rather than the actual contribution to the global objective. Under non-IID settings, prioritizing such proxies can degrade global performance. VARS-FL addresses this limitation by replacing local proxies with a direct, server-side measurement of each client's contribution via validation-loss reduction.

\subsection{Reputation and Trust Mechanisms in Federated Learning}

Several works address robustness in federated learning, particularly under adversarial settings. Krum~\cite{blanchard2017machine} and its extension Multi-Krum select updates that are closest to their neighbors in parameter space, filtering out potential Byzantine clients. Bulyan~\cite{el2018distributed} further improves robustness by combining neighbor-based filtering with coordinate-wise median aggregation. FLTrust~\cite{cao2021fltrust} introduces a server-side trusted dataset to compute a reference gradient, scoring client updates based on cosine similarity. Updates that deviate from the trusted direction are down-weighted or discarded, providing robustness against malicious participants. However, FLTrust is designed for adversarial settings and evaluates directional similarity rather than contribution to global generalization. In contrast, VARS-FL assumes honest but heterogeneous clients and focuses on measuring contribution to the global objective. It uses validation-loss reduction as a scoring signal and accumulates this signal over time into a reputation score, whereas FLTrust operates on a per-round basis without temporal aggregation. Personalized FL methods, such as FedRep~\cite{collins2021exploiting} and pFedMe~\cite{t2020personalized}, maintain per-client model components tailored to local data. While these approaches capture client-specific characteristics, they do not leverage this information for global client selection. To our knowledge, VARS-FL is the first framework to explicitly maintain a temporally aggregated, validation-aligned reputation score for client selection.

\subsection{Bandit-Based Optimisation for Client Selection}

Client selection can be naturally formulated as a multi-armed bandit (MAB) problem~\cite{lai2021oort,cho2022power}, where each client represents an arm and the objective is to balance exploration and exploitation. Oort~\cite{lai2021oort} employs an upper-confidence-bound-like strategy combining utility and system efficiency. However, existing bandit-based approaches rely on local proxy signals that are misaligned with the global objective. VARS-FL addresses this limitation by defining the reward as validation-loss reduction, aligning the bandit objective with the global training objective. The resulting reputation score $R_i^t$ can be interpreted as a non-stationary bandit estimator~\cite{garivier2011upper}, combining short-term quality estimates with long-term participation. Thompson Sampling~\cite{russo2018tutorial} provides a Bayesian alternative with strong theoretical guarantees in stationary settings. However, federated learning is inherently non-stationary, as the global model evolves over time. Prior work~\cite{garivier2011upper} shows that sliding-window or discounted approaches are necessary in such settings. VARS-FL incorporates this insight through a sliding-window quality estimator, enabling adaptation to changing client utility.

\subsection{Federated Learning for IoT Security and Intrusion Detection}

Federated learning is particularly well-suited to IoT and IIoT environments, where data is distributed, sensitive, and difficult to centralize~\cite{mothukuri2021survey}. In this context, FL enables collaborative intrusion detection without exposing raw network traffic. Several FL-based intrusion detection systems (FL-IDS) have been proposed. D\"IoT~\cite{nguyen2019diot} learns device-specific behavioral models across distributed gateways, achieving high detection performance. Rey~et~al.~\cite{rey2022federated} demonstrate that FL achieves performance comparable to centralized training while improving robustness through data diversity, although they also highlight vulnerability to poisoning attacks. Rahman~et~al.~\cite{rahman2020internet} show that FL outperforms local models and approaches centralized performance under heterogeneous data distributions. Common benchmark datasets include CICIDS2017~\cite{sharafaldin2018toward}, ToN-IoT~\cite{moustafa2021ton}, and N-BaIoT~\cite{meidan2018n}. However, these datasets have limitations in realism or scope. Edge-IIoTset~\cite{ferrag2022edge} provides a more comprehensive benchmark, with over 2.2 million samples across 15 classes, including rare attack types and significant class imbalance, making it well-suited for evaluating client selection under non-IID conditions. Despite these advances, existing FL-IDS works primarily rely on FedAvg or simple selection schemes and do not address objective misalignment. VARS-FL fills this gap by providing a validation-aligned client selection mechanism tailored to heterogeneous IoT environments.

\subsection{Summary and Positioning}

Table~\ref{tab:related} summarizes the key design dimensions of existing client selection and related federated learning methods. A clear pattern emerges: most approaches rely on \emph{local proxy signals} (e.g., local loss, gradient norms, or correlations) that are not inherently aligned with the global optimization objective, and are therefore unreliable under heterogeneous (non-IID) data distributions. In addition, existing methods are largely \emph{stateless}, treating each round independently and failing to accumulate evidence about a client’s historical contribution. While some works address system heterogeneity or adversarial robustness, none simultaneously tackle \emph{objective misalignment} and \emph{data heterogeneity}, nor do they evaluate client selection in realistic IoT and cybersecurity settings characterized by severe class imbalance and distributed data. VARS-FL addresses these limitations through a unified framework that (i) defines client utility via a \emph{globally aligned} validation-loss reduction signal, (ii) maintains a \emph{history-aware reputation score} that aggregates per-client contributions across rounds, and (iii) employs an \emph{explore--exploit selection strategy} to balance discovery and exploitation under non-stationary conditions. To the best of our knowledge, VARS-FL is the first method to combine these three properties within a single framework while remaining fully compatible with standard federated learning pipelines, thereby providing a principled and practical solution for client selection in real-world non-IID and IoT-driven environments.

\begin{figure}[t] \centering \includegraphics[width=\columnwidth]{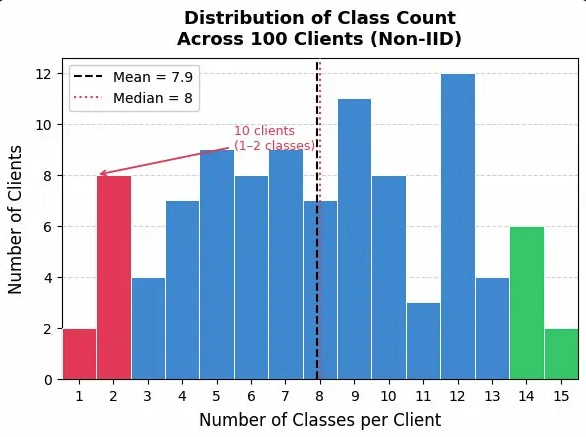} \caption{\textcolor{black}{Non-IID class presence across 100 clients (seed 42). Each row is a client sorted by class count; each column is an attack class. White indicates absence. Dataset sizes range from 426 to 5,152 samples per client (mean: 3,250); the number of local classes ranges from 1 to 15 (mean: 7.9), reflecting the heterogeneous data distributions characteristic of real IoT deployments.}} \label{fig:noniid} \end{figure}


\begin{table}[t]
\centering
\caption{Summary of Notation Used in the System Model}
\label{tab:notation}
\renewcommand{\arraystretch}{1.2}
\begin{tabular}{ll}
\toprule
\textbf{Symbol} & \textbf{Description} \\
\midrule

$\mathcal{C}$ & Set of all clients \\
$N$ & Total number of clients \\
$i$ & Client index \\
$t$ & Communication round index \\

$S_t$ & Set of selected clients at round $t$ \\
$m$ & Number of clients selected per round \\

$\mathcal{D}_i$ & Local dataset of client $i$ \\
$n_i$ & Number of samples at client $i$ \\
$\mathcal{D}_{val}$ & Server-side validation dataset \\

$\theta$ & Global model parameters \\
$\theta^{t}$ & Global model at round $t$ \\
$\theta_i^{t}$ & Local model returned by client $i$ at round $t$ \\

$E$ & Number of local training epochs \\

$\mathcal{L}_{\text{val}}(\cdot)$ & Validation loss function \\

$\delta_i^t$ & Validation-loss improvement of client $i$ \\
$Q_i^t$ & Quality score of client $i$ at round $t$ \\

$H_i$ & History of past quality scores for client $i$ \\
$\bar{Q}_i^{(t)}$ & Sliding-window average of recent quality scores \\
$R_i^t$ & Reputation score of client $i$ at round $t$ \\
$p_i^t$ & Participation count of client $i$ \\

$W$ & Sliding-window size for quality averaging \\

$\varepsilon$ & Minimum floor value for normalized quality score \\
$\zeta$ & Small constant for numerical stability \\

$\rho$ & Exploration rate in client selection \\
$T_0$ & Number of cold-start rounds \\

$w$ & Global model (algorithm notation) \\
$w_i$ & Local model returned by client $i$ (algorithm notation) \\

$C_{\mathrm{fwd}}$ & Forward-pass computational cost per sample \\
$\Delta C_{\mathrm{server}}$ & Additional server-side computation per round \\

$d_k$ & Dimension of layer $k$ in the neural network \\

$\mathbf{x}$ & Input feature vector \\
$\mathbf{h}_k$ & Hidden layer representation at layer $k$ \\
$\mathbf{z}_k$ & Post-dropout activation at layer $k$ \\
$\hat{\mathbf{y}}$ & Predicted output probability vector \\

$\mathbf{W}_k, \mathbf{b}_k$ & Weight matrix and bias vector of layer $k$ \\

\bottomrule
\end{tabular}
\end{table}

\begin{figure*}[t]
    \centering
    \includegraphics[width=\textwidth]{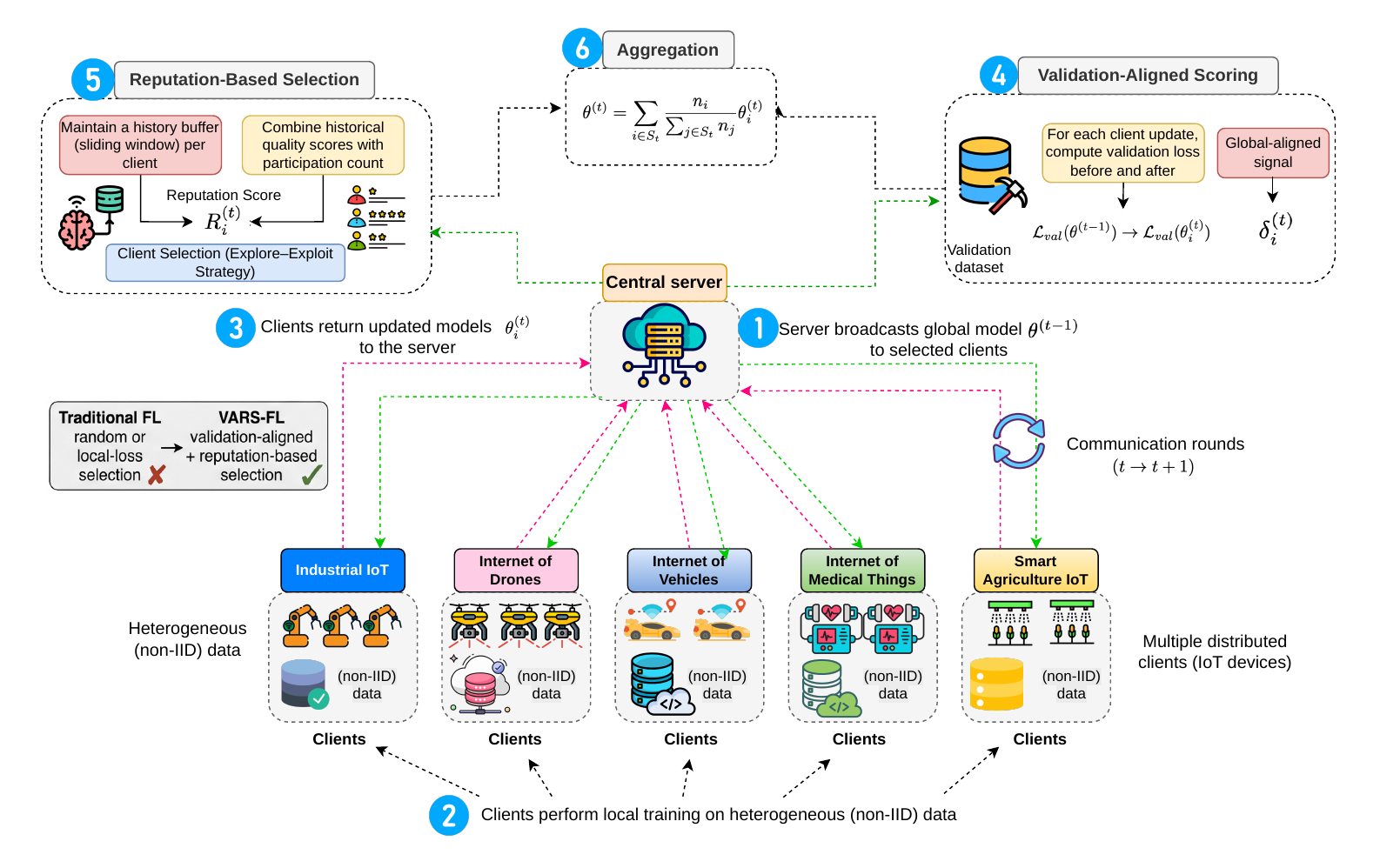}
    \caption{
    Overview of \textbf{VARS-FL}, a validation-aligned, reputation-based client selection framework for federated learning. 
    In each communication round, the server broadcasts the global model to selected clients, which perform local training on heterogeneous (non-IID) data and return updated models. 
    The server evaluates each client’s contribution using validation-loss improvement, producing a globally aligned quality signal that is aggregated over time into a reputation score. 
    Client selection is then performed using an explore–exploit strategy based on these reputation scores, while aggregation remains unchanged via FedAvg.
    }
    \label{fig:vars_fl_overview}
\end{figure*}

\section{System Model}
\label{sec:system}

Figure~\ref{fig:vars_fl_overview} provides an overview of VARS-FL, where client selection is driven by validation-aligned contribution signals and accumulated reputation, enabling objective-consistent and history-aware participation under non-IID data. Specifically, VARS-FL consists of two types of participants. The \textbf{server} coordinates the training process, maintains the global model $\theta$, holds a shared validation set, and tracks all client-level statistics. The \textbf{clients} ($\mathcal{C} = \{1, \ldots, N\}$) retain their local datasets $\mathcal{D}_i$ and perform training locally, returning updated model parameters to the server without exposing raw data. Table~\ref{tab:notation} summarizes the notation used throughout this section, providing definitions for all variables, parameters, and symbols employed in the formulation of the VARS-FL framework.

Each communication round $t$ proceeds as follows:
\begin{enumerate}
    \item The server selects a subset $S_t \subseteq \mathcal{C}$ with $|S_t| = m$ and broadcasts the current global model $\theta^{t-1}$.
    \item Each client $i \in S_t$ trains locally on $\mathcal{D}_i$ for $E$ epochs and returns updated parameters $\theta_i^t$.
    \item The server evaluates each client's contribution via a quality score $Q_i^t$ and updates its Reputation score $R_i^t$.
    \item The server aggregates the updates using standard FedAvg:
\end{enumerate}

\begin{equation}
\theta^{t} = \sum_{i \in S_t} \frac{n_i}{\sum_{j \in S_t} n_j} \theta_i^t,
\label{eq:fedavg}
\end{equation}

where $n_i$ denotes the number of local samples held by client $i$. Reputation scores influence \emph{client selection} in subsequent rounds but are not used as aggregation weights, ensuring full compatibility with the standard FedAvg protocol.

\subsection{Data Preprocessing}
\label{sec:preprocessing}

The Edge-IIoTset dataset contains 63 raw features, of which 19 correspond to non-numeric protocol fields---including IP address strings, HTTP URI components, MQTT payload content, and TCP port identifiers---that lack a meaningful ordinal structure for gradient-based learning. These features are removed, resulting in a 43-dimensional numeric feature space. The retained features capture packet-level statistics, flow-level counters, and protocol-specific attributes across TCP, UDP, DNS, HTTP, MQTT, and ARP layers. No further feature selection or dimensionality reduction is applied in order to preserve the full set of available numerical signals.

To mitigate class imbalance, the majority class (\textit{Normal}) is capped at 18\% of the total sample count. In the original dataset, Normal traffic accounts for 1,615,643 out of 2,219,201 samples (72.8\%); after capping, it is reduced to 132,488 samples, yielding a balanced dataset of 736,046 samples across 15 classes. This strategy prevents the model from collapsing to a majority-class predictor while preserving a realistic distribution of attack categories. All features are standardized to zero mean and unit variance using statistics computed on the training set, and the same transformation is applied to validation and test splits to avoid data leakage.

The dataset is partitioned into training, validation, and test sets using a 70/15/15 split, corresponding to 515,232, 110,407, and 110,407 samples, respectively. The validation set is maintained exclusively at the server and is used solely for client quality evaluation (Eq.~\ref{eq:delta}); it is neither shared with clients nor used for gradient updates. Training data is distributed across clients using a heterogeneous non-IID partitioning scheme. Client dataset sizes range from 426 to 5,152 samples (mean: 3,250), and the number of classes per client varies from 1 to 15 (mean: 7.9, median: 8). This results in a highly heterogeneous distribution, where some clients observe only a single attack type while others cover the full class space, reflecting the diversity of real-world IoT deployments. The resulting non-IID structure is illustrated in Fig.~\ref{fig:noniid}.


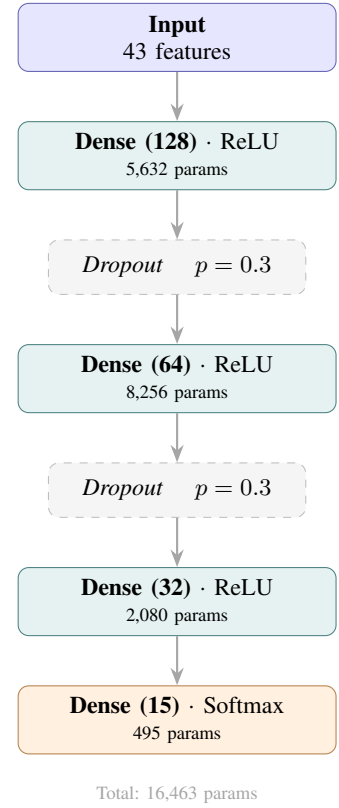
\begin{figure}[h]
\centering
\begin{tikzpicture}[
    node distance=0.65cm,
    >=Latex,
    every node/.style={font=\small},
    layer/.style={
        draw,
        rounded corners=4pt,
        minimum width=4.2cm,
        minimum height=0.9cm,
        align=center,
        line width=0.4pt
    },
    input/.style={layer,
        fill=blue!10, draw=blue!50!black!60},
    hidden/.style={layer,
        fill=teal!10, draw=teal!60!black!70},
    dropout/.style={layer,
        fill=gray!8, draw=gray!50,
        dashed, minimum height=0.72cm,
        minimum width=3.4cm, font=\small\itshape},
    output/.style={layer,
        fill=orange!12, draw=orange!60!black!70},
    arrow/.style={->, >=Stealth, thick, color=gray!70},
    param/.style={font=\scriptsize, text=gray!70}
]

\node[input]   (in)  {\textbf{Input} \\ \small 43 features};
\node[hidden]  (h1)  [below=of in]
    {\textbf{Dense (128)} $\cdot$ ReLU \\ \scriptsize 5{,}632 params};
\node[dropout] (d1)  [below=of h1]   {Dropout \quad $p = 0.3$};
\node[hidden]  (h2)  [below=of d1]
    {\textbf{Dense (64)} $\cdot$ ReLU \\ \scriptsize 8{,}256 params};
\node[dropout] (d2)  [below=of h2]   {Dropout \quad $p = 0.3$};
\node[hidden]  (h3)  [below=of d2]
    {\textbf{Dense (32)} $\cdot$ ReLU \\ \scriptsize 2{,}080 params};
\node[output]  (out) [below=of h3]
    {\textbf{Dense (15)} $\cdot$ Softmax \\ \scriptsize 495 params};

\draw[arrow] (in)  -- (h1);
\draw[arrow] (h1)  -- (d1);
\draw[arrow] (d1)  -- (h2);
\draw[arrow] (h2)  -- (d2);
\draw[arrow] (d2)  -- (h3);
\draw[arrow] (h3)  -- (out);

\node[param, below=0.3cm of out]
    {Total: 16{,}463 params};

\end{tikzpicture}
\caption{\textcolor{black}{Architecture of the fully connected DNN used in the federated learning experiments. The model takes 43 standardized numeric features as input, followed by three hidden layers of sizes 128, 64, and 32 with ReLU activations. Dropout ($p=0.3$) is applied after the first two hidden layers, and the output layer uses softmax activation for multi-class classification.}}
\label{fig:dnn_architecture}
\end{figure}

\subsection{Model Architecture}
\label{sec:architecture}

VARS-FL is model-agnostic by design: the server-side quality scoring mechanism evaluates the validation loss of any model returned by clients, independent of architecture or parameterization. In this work, we instantiate VARS-FL using a fully connected deep neural network (DNN), a compact and interpretable architecture well-suited for tabular network traffic data.

\textbf{Input representation.} Each sample is represented as a 43-dimensional real-valued vector $\mathbf{x} \in \mathbb{R}^{43}$, corresponding to the standardized numeric features.

\textbf{Hidden layers.} The network consists of three fully connected layers with decreasing width and ReLU activations:
\begin{align}
\mathbf{h}_1 &= \mathrm{ReLU}(\mathbf{W}_1 \mathbf{x} + \mathbf{b}_1), 
\quad \mathbf{W}_1 \in \mathbb{R}^{128 \times 43}, \\
\mathbf{h}_2 &= \mathrm{ReLU}(\mathbf{W}_2 \mathbf{z}_1 + \mathbf{b}_2), 
\quad \mathbf{W}_2 \in \mathbb{R}^{64 \times 128}, \\
\mathbf{h}_3 &= \mathrm{ReLU}(\mathbf{W}_3 \mathbf{z}_2 + \mathbf{b}_3), 
\quad \mathbf{W}_3 \in \mathbb{R}^{32 \times 64},
\end{align}
where $\mathbf{z}_1 = \mathrm{Dropout}(\mathbf{h}_1, p=0.3)$ and $\mathbf{z}_2 = \mathrm{Dropout}(\mathbf{h}_2, p=0.3)$. Dropout is applied during training to mitigate overfitting under heterogeneous client data distributions and is disabled at inference time.

\textbf{Output layer.} A linear projection followed by softmax produces class probabilities over the 15 traffic classes:
\begin{equation}
\hat{\mathbf{y}} = \mathrm{softmax}(\mathbf{W}_4 \mathbf{h}_3 + \mathbf{b}_4), 
\quad \mathbf{W}_4 \in \mathbb{R}^{15 \times 32}.
\end{equation}

\textbf{Model size.} The total number of trainable parameters is:
\begin{align}
|\theta| &= (43 \cdot 128 + 128) + (128 \cdot 64 + 64) \nonumber \\
         &\quad + (64 \cdot 32 + 32) + (32 \cdot 15 + 15) \nonumber \\
         &= 16{,}463.
\end{align}
This compact design reduces communication overhead, which is critical in bandwidth-constrained federated environments.

\textbf{Training configuration.} The model is trained using the Adam optimizer with learning rate $\eta = 10^{-3}$ and categorical cross-entropy loss. Federated hyperparameters are summarized in Table~\ref{tab:settings}.

\subsection{Validation-Loss Improvement}

At the end of each round, the server assigns a quality score $Q_i^t \in [0,1]$ to each participating client, reflecting its contribution to global model improvement as measured on the server-side validation set.

To compute this signal, the server evaluates the validation loss before and after applying each client’s update. The validation-loss improvement is defined as:
\begin{equation}
\delta_i^t = \max\!\left(0,\;
\mathcal{L}_{\text{val}}(\theta^{t-1}) -
\mathcal{L}_{\text{val}}(\theta_i^t)\right),
\label{eq:delta}
\end{equation}
where $\theta_i^t$ denotes the model obtained by applying client $i$’s update. Updates that degrade validation performance receive $\delta_i^t = 0$.

To enable relative comparison within a round, the scores are normalized:
\begin{equation}
Q_i^t = \max\!\left(\varepsilon,\;
\frac{\delta_i^t}{\max_{j \in S_t} \delta_j^t + \zeta}\right),
\label{eq:quality}
\end{equation}
where $\varepsilon > 0$ is a small floor value and $\zeta$ ensures numerical stability. This normalization emphasizes relative contribution and prevents dominance by a single client, while $\varepsilon$ ensures continued exploration.

\begin{algorithm}[t]
\caption{VARS-FL Training Procedure}
\label{alg:varsfl-main}
\begin{algorithmic}[1]
\REQUIRE Clients $\mathcal{C}=\{1,\dots,N\}$, rounds $T$, clients per round $m$, validation set $\mathcal{D}_{val}$, initial model $w_0$
\STATE Initialize global model $w \leftarrow w_0$
\FORALL{clients $i \in \mathcal{C}$}
    \STATE Initialize participation count $p_i \leftarrow 0$
    \STATE Initialize quality history $H_i \leftarrow [\,]$
\ENDFOR
\FOR{$t=1$ \TO $T$}
    \STATE Select clients $S_t$ using Algorithm~\ref{alg:varsfl-selection}
    \FORALL{clients $i \in S_t$}
        \STATE Send $w$ to client $i$
        \STATE Client trains locally and returns updated model $w_i$ and sample count $n_i$
    \ENDFOR
    \STATE Update client quality histories and participation counts using Algorithm~\ref{alg:varsfl-scoring}
    \STATE Aggregate updates using FedAvg:
    \STATE $w \leftarrow \sum_{i\in S_t}\frac{n_i}{\sum_{j\in S_t} n_j}\, w_i$
\ENDFOR
\RETURN $w$
\end{algorithmic}
\end{algorithm}

\subsection{From Quality Scores to Reputation}

Single-round quality scores can be noisy; therefore, VARS-FL aggregates evidence over time through a \emph{Reputation} score. For each client, we maintain a sliding-window average $\bar{Q}_i^{(t)}$ over the most recent $W$ scores and a participation count $p_i^t$. The reputation is defined as:
\begin{equation}
R_i^t = \bar{Q}_i^{(t)} \cdot \log(1 + p_i^t),
\label{eq:Reputation}
\end{equation}
where $W=5$ in our experiments. The logarithmic term provides diminishing returns for frequent participation, ensuring that consistent quality---rather than frequency alone---drives selection priority.

\subsection{Explore--Exploit Selection}
\label{sec:bandit}

Client selection is formulated as a multi-armed bandit (MAB) problem, where each client represents an arm and rewards are observed only upon selection. In VARS-FL, the reward corresponds to the validation-aligned quality score $Q_i^t$, and the objective is to maximize cumulative validation-loss reduction under a budget of $m$ clients per round.

We adopt a hybrid explore--exploit strategy: a majority of clients are selected based on the highest reputation scores (exploitation), while the remaining slots are filled via uniform random sampling (exploration). This approach ensures that potentially informative clients are not permanently excluded, while prioritizing those with demonstrated utility. The resulting policy is robust to both non-IID data and non-stationary client contributions.

The overall VARS-FL procedure is decomposed into three components: (i) the main federated training loop (Algorithm~\ref{alg:varsfl-main}), (ii) reputation-based client selection (Algorithm~\ref{alg:varsfl-selection}), and (iii) validation-aligned quality scoring (Algorithm~\ref{alg:varsfl-scoring}). This modular decomposition highlights the separation between standard federated optimization and the proposed selection and scoring mechanisms.

\begin{algorithm}[t]
\caption{Reputation-Based Client Selection}
\label{alg:varsfl-selection}
\begin{algorithmic}[1]
\REQUIRE Clients $\mathcal{C}$, histories $\{H_i\}$, participation counts $\{p_i\}$, clients per round $m$, exploration rate $\rho$, cold-start rounds $T_0$, current round $t$
\FORALL{clients $i \in \mathcal{C}$}
    \IF{$|H_i|>0$}
        \STATE $\bar{Q}_i \leftarrow \frac{1}{|H_i|}\sum_{q\in H_i} q$
        \STATE $R_i \leftarrow \bar{Q}_i \cdot \log(1+p_i)$
    \ELSE
        \STATE $R_i \leftarrow 0$
    \ENDIF
\ENDFOR
\IF{$t \le T_0$}
    \STATE Sample $S_t \subset \mathcal{C}$ uniformly at random such that $|S_t|=m$
\ELSE
    \STATE $m_{\mathrm{rep}} \leftarrow \lfloor(1-\rho)m\rfloor$
    \STATE $m_{\mathrm{rnd}} \leftarrow m - m_{\mathrm{rep}}$
    \STATE $S_t^{\mathrm{rep}} \leftarrow$ top-$m_{\mathrm{rep}}$ clients according to $R_i$
    \STATE Sample $S_t^{\mathrm{rnd}} \subset \mathcal{C}\setminus S_t^{\mathrm{rep}}$ uniformly at random such that $|S_t^{\mathrm{rnd}}|=m_{\mathrm{rnd}}$
    \STATE $S_t \leftarrow S_t^{\mathrm{rep}} \cup S_t^{\mathrm{rnd}}$
\ENDIF
\RETURN $S_t$
\end{algorithmic}
\end{algorithm}

\begin{algorithm}[t]
\caption{Validation-Aligned Quality Scoring}
\label{alg:varsfl-scoring}
\begin{algorithmic}[1]
\REQUIRE Selected clients $S_t$, global model $w$, returned models $\{w_i\}_{i\in S_t}$, validation set $\mathcal{D}_{val}$, histories $\{H_i\}$, participation counts $\{p_i\}$, window size $W$
\STATE $L_{\mathrm{base}} \leftarrow \mathcal{L}_{val}(w;\mathcal{D}_{val})$
\FORALL{clients $i \in S_t$}
    \STATE $L_i \leftarrow \mathcal{L}_{val}(w_i;\mathcal{D}_{val})$
    \STATE $\delta_i \leftarrow \max(0, L_{\mathrm{base}} - L_i)$
\ENDFOR
\STATE $\delta_{\max} \leftarrow \max_{j\in S_t}\delta_j$
\FORALL{clients $i \in S_t$}
    \STATE $Q_i \leftarrow \max\left(\varepsilon,\frac{\delta_i}{\delta_{\max}+\zeta}\right)$
    \STATE Append $Q_i$ to $H_i$
    \IF{$|H_i|>W$}
        \STATE Remove the oldest element from $H_i$
    \ENDIF
    \STATE $p_i \leftarrow p_i + 1$
\ENDFOR
\RETURN Updated $\{H_i\}$ and $\{p_i\}$
\end{algorithmic}
\end{algorithm}

\subsection{Computational and Communication Complexity}
\label{sec:complexity}

\textbf{Communication overhead.}
In each round, every selected client uploads a full copy of the model parameters $\theta_i^t$ to the server. For the DNN used in this work, this corresponds to $|\theta| = 16{,}463$ parameters, i.e., $16{,}463 \times 32 = 526{,}816$ bits ($\approx$64~KB) per client per round under 32-bit floating-point representation. For $m$ participating clients, the total uplink communication per round is therefore $m \cdot 64$~KB.

Importantly, VARS-FL introduces \emph{no additional communication overhead} beyond standard FedAvg (Algorithm~\ref{alg:varsfl-main}). Clients transmit only model updates and do not report any auxiliary statistics. All additional computations associated with client selection and scoring (Algorithms~\ref{alg:varsfl-selection} and~\ref{alg:varsfl-scoring}) are performed entirely on the server. Consequently, VARS-FL maintains identical communication cost to FedAvg and incurs strictly lower overhead than methods that require extra client-side information, such as local loss (Oort, Power-of-Choice) or full gradients (FedCor).

\textbf{Server-side overhead.}
The additional computational cost of VARS-FL arises from validation-based client scoring (Algorithm~\ref{alg:varsfl-scoring}). For each selected client, the server performs a forward pass of the returned model $\theta_i^t$ over the validation set $\mathcal{D}_{val}$ to compute $\mathcal{L}_{\text{val}}(\theta_i^t)$. This operation involves only inference (no gradient computation) and is fully parallelizable across clients.

The per-sample forward-pass cost of the DNN is:
\begin{equation}
\begin{aligned}
C_{\mathrm{fwd}}
&= \sum_k d_{k-1} d_k \\
&= 43\cdot128 + 128\cdot64 + 64\cdot32 + 32\cdot15 \\
&= 16{,}224\ \text{MACs},
\end{aligned}
\end{equation}
where MAC denotes a multiply-accumulate operation and $d_k$ is the dimension of layer $k$. The total additional server-side cost per round is:
\begin{equation}
\Delta C_{\mathrm{server}} = m \cdot |\mathcal{D}_{val}| \cdot C_{\mathrm{fwd}}.
\end{equation}

For our configuration ($m=10$, $|\mathcal{D}_{val}|=110{,}407$), this yields approximately $1.79 \times 10^{10}$ MACs per round. Since inference is significantly cheaper than training and can be parallelized, this overhead remains modest compared to the cumulative cost of local training across clients.

\textbf{Scalability.}
A key property of VARS-FL is that its additional computational cost depends only on the number of selected clients $m$ and the validation set size $|\mathcal{D}_{val}|$, but is \emph{independent of the total number of clients $N$}. As a result, the overhead remains constant as the federation scales, making VARS-FL well-suited for large-scale deployments. This contrasts with methods such as FedCor, whose complexity scales as $\mathcal{O}(m^2 \cdot |\theta|)$ due to pairwise client interactions.

\textbf{Memory overhead.}
VARS-FL maintains a history buffer of size $W$ for each client, resulting in a total storage cost of $\mathcal{O}(N \cdot W)$ scalar values on the server (Algorithm~\ref{alg:varsfl-selection}). In practice, this overhead is negligible; for $N=100$ and $W=5$, it corresponds to only 500 floating-point values.

Table~\ref{tab:complexity} summarizes the communication and computational complexity of all compared methods.

\begin{table*}[t]
\centering
\caption{Complexity comparison of client selection methods}
\label{tab:complexity}
\renewcommand{\arraystretch}{1.2}
\begin{tabular}{lccc}
\toprule
\textbf{Method}
  & \textbf{Extra Client Comm.}
  & \textbf{Server Cost}
  & \textbf{State} \\
\midrule
FedAvg~\cite{mcmahan2017communication}
  & None
  & None
  & None \\
PoC~\cite{cho2022power}
  & Loss scalar
  & $\mathcal{O}(1)$ per client
  & None \\
Oort~\cite{lai2021oort}
  & Loss scalar
  & $\mathcal{O}(1)$ per client
  & None \\
FedCor~\cite{tang2022fedcor}
  & Full gradient ($|\theta|$)
  & $\mathcal{O}(m^2 \cdot |\theta|)$
  & None \\
\midrule
\textbf{VARS-FL (Ours)}
  & \textbf{None}
  & $\mathcal{O}(m \cdot |\mathcal{D}_{val}| \cdot C_{\mathrm{fwd}})$
  & $\mathcal{O}(N \cdot W)$ \\
\bottomrule
\end{tabular}
\end{table*}

\begin{figure}[h]
  \centering
  \includegraphics[width=\columnwidth]{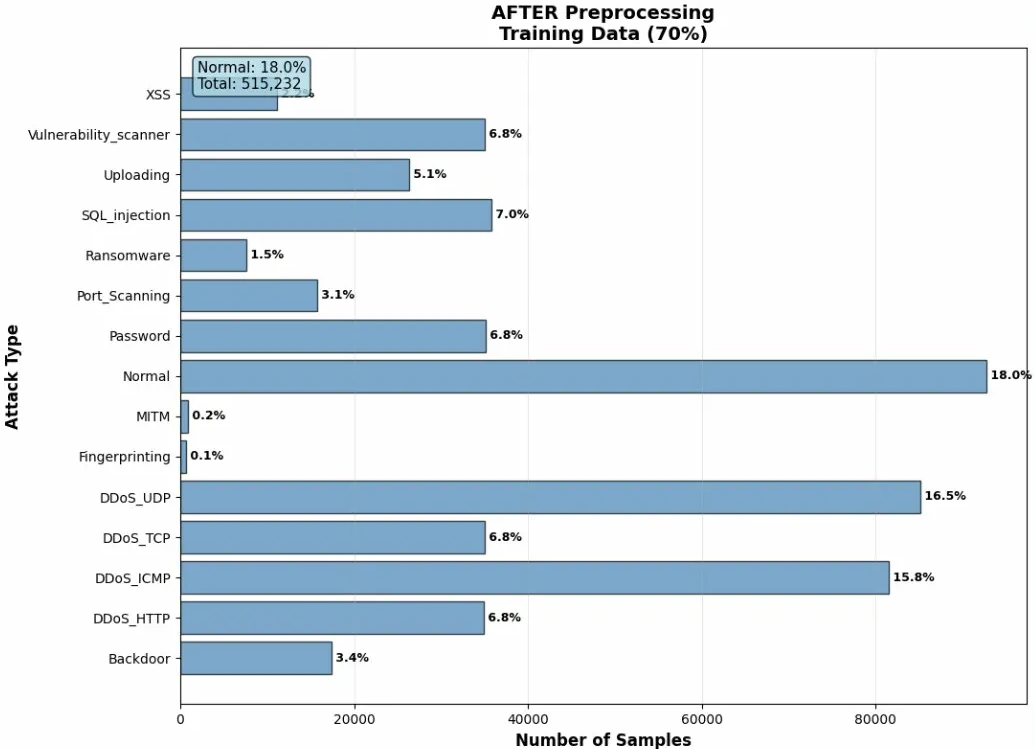}
  \caption{Class distribution of the training split (515,232 samples)
  after preprocessing. Normal is capped at 18\% to prevent
  majority-class dominance. MITM (0.2\%), Fingerprinting (0.1\%),
  and Ransomware (1.5\%) are the rarest classes; validation and test
  splits share identical proportions, confirming the server validation
  set is representative of the test distribution.}
  \label{fig:dataset}
\end{figure}

\section{Experimental Setup}
\label{sec:setup}

\subsection{Dataset and Non-IID Data Partitioning}

We evaluate VARS-FL on the Edge-IIoTset Cyber Security Dataset~\cite{ferrag2022edge}, a large-scale and realistic IoT/IIoT network traffic benchmark comprising 2,219,201 samples with 63 raw features and 15 traffic classes. These include one benign class (\textit{Normal}) and 14 attack categories, such as Ransomware, SQL injection, multiple DDoS variants (HTTP, TCP, UDP, ICMP), MITM, XSS, Backdoor, Port Scanning, Vulnerability Scanner, Uploading, Fingerprinting, and Password attacks.

The dataset is partitioned into training, validation, and test sets using a 70/15/15 split, corresponding to 515,232, 110,407, and 110,407 samples, respectively. The validation set is maintained exclusively at the server and is used solely for client quality evaluation via $\delta_i^t$ (Eq.~\ref{eq:delta}); it is not used for model training.

To reflect realistic deployment conditions, the training data is distributed across clients using a heterogeneous non-IID partitioning scheme. This setup captures the variability inherent in IoT environments, where individual nodes observe different traffic patterns and attack types. In the 100-client setting, local dataset sizes range from 426 to 5,152 samples (mean: 3,250), and the number of classes per client varies from 1 to 15. This results in a highly skewed distribution, where some clients contain only a single attack type while others observe a broader spectrum of traffic, as illustrated in Fig.~\ref{fig:dataset}.

\subsection{Models and Configurations}

We compare VARS-FL against three representative client selection baselines: \emph{FedAvg}~\cite{mcmahan2017communication}, \emph{Oort}~\cite{lai2021oort}, and \emph{Power-of-Choice}~\cite{cho2022power}. To ensure a fair comparison, all methods use the same model architecture, optimizer, learning rate, number of local epochs, and aggregation rule (FedAvg). This isolates the effect of client selection from other factors.

The experimental configuration is summarized in Table~\ref{tab:settings}. We run all experiments for $T=100$ communication rounds with a participation rate of $m=10\%$ per round. Local training is performed for $E=3$ epochs using the Adam optimizer with learning rate $\eta=10^{-3}$ and batch size 256. To evaluate robustness, results are averaged across three random seeds (7, 42, and 123). VARS-FL-specific hyperparameters are set to $T_0=15$ (cold-start rounds), $\rho=0.3$ (exploration rate), and $W=5$ (reputation window size).

\begin{table}[t]
\caption{Experimental configuration}
\label{tab:settings}
\centering
\small
\begin{tabular}{ll}
\toprule
\textbf{Item} & \textbf{Setting} \\
\midrule
Rounds & $T=100$ \\
Clients per round & $m=10\%$ \\
Local epochs & $E=3$ \\
Optimizer & Adam \\
Learning rate & $\eta=10^{-3}$ \\
Batch size & 256 \\
Baselines & FedAvg, Oort, Power-of-Choice \\
Seeds (robustness) & 7, 42, 123 \\
VARS-FL hyperparameters & $T_0=15$, $\rho=0.3$, $W=5$ \\
\bottomrule
\end{tabular}
\end{table}

\subsection{Evaluation Metrics}

We evaluate performance using test-set accuracy, macro-averaged F1 score (F1-Macro), weighted F1 score, and cross-entropy loss. Among these, F1-Macro is the primary metric, as it assigns equal importance to all classes and therefore provides a more informative assessment under class imbalance. This is particularly important in intrusion detection scenarios, where rare attack types must be detected reliably despite their low frequency.

\section{Results}
\label{sec:results}

\subsection{Overall Performance}
This section evaluates whether VARS-FL provides measurable gains over established client-selection strategies. We run experiments with $N=100$ clients over $T=100$ communication rounds, comparing VARS-FL against FedAvg (baseline), Oort, and Power-of-Choice, while keeping the model architecture, optimizer, and FedAvg aggregation fixed across methods. Evaluation is done on a held-out test set using accuracy, F1-Macro, weighted F1, and cross-entropy loss, with experiments repeated across multiple random seeds to assess both average performance and robustness.

Table~\ref{tab:allseeds_meanstd} aggregates results over three random seeds(7, 42, and 123) and shows that VARS-FL achieves the best mean performance on every metric while also having strong consistency across runs. VARS-FL attains the highest mean Accuracy ($0.8185 \pm 0.0040$), F1-Macro ($0.6422 \pm 0.0241$), F1-Weighted ($0.8086 \pm 0.0080$), and Precision ($0.8026 \pm 0.0392$), and the lowest mean Loss ($0.4937 \pm 0.0200$). Relative to FedAvg, VARS-FL delivers consistent gains of $+0.0514$ Accuracy, $+0.0857$ F1-Macro, and $+0.0584$ F1-Weighted, while reducing Loss by $-0.0916$. Notably, VARS-FL's very small standard deviations for Accuracy and F1-Weighted indicate that its advantage is not driven by a single favorable initialization, but persists reliably across seeds.

Looking beyond average performance gains, the observed pattern of improvement reveals where VARS-FL provides the most practical value. The most significant improvement is observed in {F1-Macro}, which is sensitive to class imbalance and minority class behavior; the $+0.0857$ mean increase over FedAvg suggests VARS-FL improves class balanced generalization rather than simply optimizing dominant class accuracy. In parallel, the substantial Loss reduction implies more stable convergence and better validation-aligned updates, consistent with selecting clients whose contributions generalize rather than overfit. In comparison, Oort shows higher variability (e.g., Accuracy $0.6866$ and Loss $0.8118$), indicating sensitivity to sampling dynamics and seed effects, whereas VARS-FL maintains competitive Precision while improving both F1 scores, demonstrating a balanced reduction in errors rather than a one sided shift in the precision-recall trade-off.

\begin{table*}[t]
\centering
\caption{All Methods Aggregated Over 3 Seeds (Mean $\pm$ Std), 100 Clients, 100 Rounds}
\label{tab:allseeds_meanstd}
\setlength{\tabcolsep}{3pt}
\begin{tabular}{lccccc}
\toprule
\textbf{Metric} & \textbf{FedAvg} & \textbf{Oort} & \textbf{PoC} & \textbf{VARS-FL} & \textbf{$\Delta$ (VARS--FedAvg)} \\
\midrule
Accuracy            & $0.7671 \pm 0.0422$ & $0.6866 \pm 0.1043$ & $0.7925 \pm 0.0204$ & $\mathbf{0.8185 \pm 0.0040}$ & $+0.0514 \pm 0.0382$ \\
F1-Macro            & $0.5565 \pm 0.0699$ & $0.5183 \pm 0.0800$ & $0.5928 \pm 0.0446$ & $\mathbf{0.6422 \pm 0.0241}$ & $+0.0857 \pm 0.0525$ \\
F1-Weighted         & $0.7502 \pm 0.0442$ & $0.6799 \pm 0.0894$ & $0.7745 \pm 0.0301$ & $\mathbf{0.8086 \pm 0.0080}$ & $+0.0584 \pm 0.0369$ \\
Precision           & $0.7281 \pm 0.0365$ & $0.7160 \pm 0.0706$ & $0.7544 \pm 0.0275$ & $\mathbf{0.8026 \pm 0.0392}$ & $+0.0745 \pm 0.0227$ \\
Loss ($\downarrow$) & $0.5852 \pm 0.0657$ & $0.8118 \pm 0.2257$ & $0.5911 \pm 0.0744$ & $\mathbf{0.4937 \pm 0.0200}$ & $-0.0916 \pm 0.0457$ \\
\bottomrule
\end{tabular}
\end{table*}

\subsection{Convergence Speed}
Table~\ref{tab:convergence} shows that VARS-FL converges significantly faster and more reliably than FedAvg, Oort, and PoC across all three seeds, as measured by rounds required to reach fixed accuracy thresholds. At the 75\% threshold, VARS-FL is consistently the fastest method, requiring only 16, 20, and 18 rounds for Seeds 7, 42, and 123, corresponding to approximately $\times$1.9, $\times$1.4, and $\times$1.8 speedups versus FedAvg, respectively; it also outpaces PoC in every case (e.g., 16 vs.\ 20 rounds in Seed~7).

At the 80\% threshold, VARS-FL is the only method that \emph{consistently reaches the target} across all seeds, achieving 63, 60, and 58 rounds, while baselines frequently fail to reach 80\% within 100 rounds (FedAvg fails in Seeds 42 and 123; Oort fails in all three; PoC reaches the threshold but requires more rounds, e.g. 93 vs.\ 60 in Seed~42 and 67vs.\ 58 in Seed~123). 

Overall, these results indicate that VARS-FL not only improves final performance but also accelerates time-to-accuracy and increases robustness to seed variability, particularly at higher-accuracy thresholds where weaker selection strategies tend to stall.

\begin{table}[H]
\centering
\caption{Rounds to Reach Accuracy Threshold (DNN, 100 Clients)}
\label{tab:convergence}
\setlength{\tabcolsep}{3pt}
\begin{tabular}{lcccccc}
\toprule
\textbf{Seed} & \textbf{Thresh.} & \textbf{FedAvg} &
\textbf{Oort} & \textbf{PoC} & \textbf{VARS-FL} & \textbf{Speedup} \\
\midrule
\multirow{2}{*}{7}
 & 75\% & 30  & 26  & 20 & \textbf{16} & $\times$1.9 vs.\ FedAvg \\
 & 80\% & 99  & --- & 99 & \textbf{63} & 36\% fewer rounds \\
\midrule
\multirow{2}{*}{42}
 & 75\% & 27  & 38  & 24 & \textbf{20} & $\times$1.4 vs.\ FedAvg \\
 & 80\% & --- & --- & 93 & \textbf{60} & FedAvg never reaches \\
\midrule
\multirow{2}{*}{123}
 & 75\% & 32  & 22  & 29 & \textbf{18} & $\times$1.8 vs.\ FedAvg \\
 & 80\% & --- & --- & 67 & \textbf{58} & FedAvg/Oort never reach \\
\bottomrule
\end{tabular}
\end{table}

\subsection{Training Dynamics}
\begin{figure}[h]
  \centering
  \includegraphics[width=0.5\textwidth]{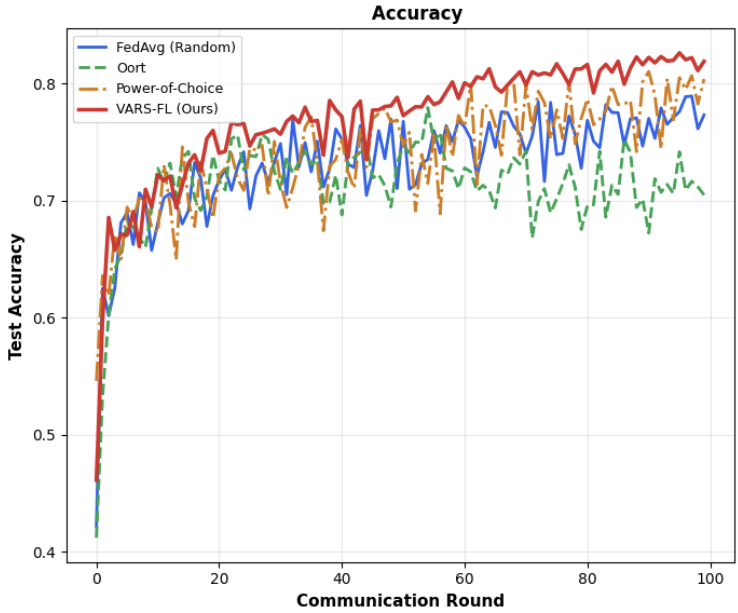}
  \caption{Test accuracy of all strategies over 100 rounds (100 clients, seed 123)}
  \label{fig:acc_123}
\end{figure}

Figure~\ref{fig:acc_123} shows the test accuracy trajectory across 100 rounds.
All methods converge at a similar rate during the cold-start phase (rounds 1--15), after which VARS-FL's reputation-driven selection takes effect and its accuracy curve separates clearly from the rest.
VARS-FL reaches the 80\% threshold at round~58, while Power-of-Choice trails at round~67 and both FedAvg and Oort fail to reach this threshold within the 100-round budget (Table~\ref{tab:convergence}).
Oort exhibits persistent high-amplitude oscillations throughout training, reflecting the instability induced by selecting clients based on local loss under this non-IID partition---rounds where locally-lossy clients are selected pull the global model in directions misaligned with the global objective, causing accuracy to fluctuate rather than converge.

\begin{figure}[t]
  \centering
  \includegraphics[width=\columnwidth]{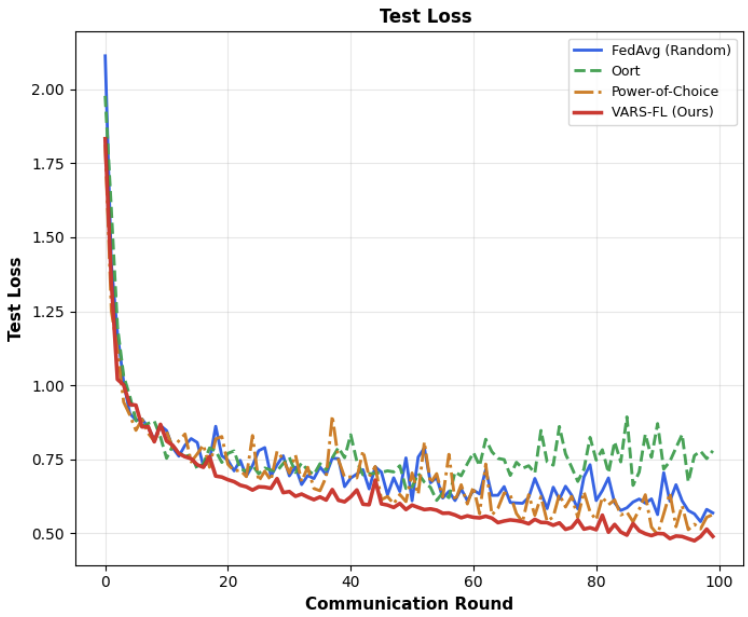}
  \caption{Test loss of all strategies over 100 rounds (100 clients, seed 123)}
  \label{fig:loss_123}
\end{figure}

\begin{figure}[t]
  \centering
  \includegraphics[width=\columnwidth]{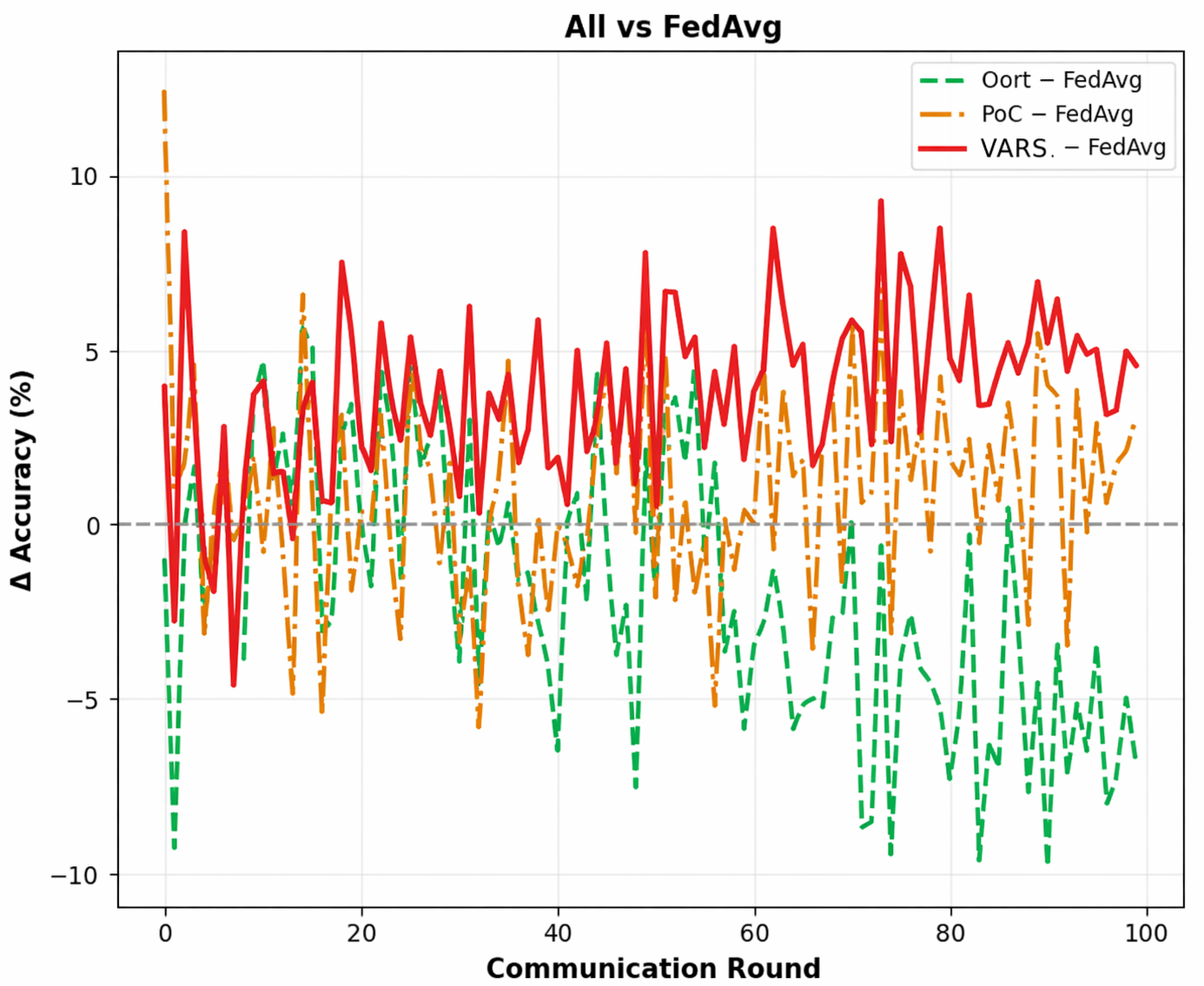}
  \caption{Per round accuracy delta relative to FedAvg for Oort, PoC, and VARS-FL (100 clients, seed 123)}
  \label{fig:delta_123}
\end{figure}

Figure~\ref{fig:loss_123} provides the clearest evidence for the objective-alignment argument.
From approximately round~20 onward, VARS-FL achieves strictly lower test loss than all baselines, converging to $0.489$ by round~100 compared to $0.568$ for FedAvg, $0.563$ for Power-of-Choice, and $0.778$ for Oort.
Oort's loss stagnates well above FedAvg for the majority of training, confirming that biasing selection toward high local-loss clients not only fails to reduce global loss but can actively impede it.
Figure~\ref{fig:delta_123} reinforces this conclusion from a relative perspective: VARS-FL's per-round accuracy delta over FedAvg (red curve) remains predominantly positive across all 100 rounds, demonstrating a sustained rather than episodic advantage.
By contrast, Oort's delta (green) oscillates around zero and trends negative after round~50, indicating that under this seed its local-loss-biased policy provides no reliable improvement over uniform random selection.

\subsection{Per-class recall analysis}
Table~\ref{tab:perclass_recall} shows that VARS-FL's gains are
concentrated where they matter most: the six classes with FedAvg
recall below 0.50 all improve under VARS-FL by $+$0.056 to
$+$0.519~pp, because validation-loss scoring preferentially
re-selects the minority-class clients that random sampling reaches
too infrequently. Oort achieves even higher recall on some of these
classes (e.g.\ Uploading~0.844) but at catastrophic cost to others
(Backdoor: 0.000, DDoS\_TCP: 0.156), whereas VARS-FL's global
reward signal prevents such single-class collapse. VARS-FL's only
meaningful regression is Backdoor ($-$0.110), attributable to
clients holding already-well-classified Backdoor data producing
small $\delta_i^t$ and thus accumulating lower Reputation scores; eight
classes remain effectively unchanged ($|\Delta| \leq 0.05$).
Overall, VARS-FL's macro-averaged recall exceeds all baselines
across every seed (Table~\ref{tab:allseeds_meanstd}), confirming that
objective-aligned selection distributes recall improvements broadly
rather than concentrating gains on a narrow class subset.
In particular, the per-class analysis in Table~\ref{tab:perclass_recall} shows that VARS-FL strengthens recall on several difficult-to-detect attack types where benchmark selection methods can either underperform or even collapse (e.g., rare or highly heterogeneous classes), indicating that validation-aligned scoring can help surface security attacks that may otherwise be missed.

\begin{table}
\centering
\caption{Per-Class Recall: All Methods — 100 Clients, 100 Rounds, Seed 42}
\label{tab:perclass_recall}
\setlength{\tabcolsep}{4pt}
\renewcommand{\arraystretch}{1.05}
\begin{tabular}{lcccc>{\columncolor{blue!5}}c}
\toprule
\textbf{Attack Type} & \textbf{FedAvg} & \textbf{Oort} & \textbf{PoC} & \textbf{VARS-FL} & \textbf{$\Delta$} \\
\midrule
Port\_Scanning        & 0.4059 & 0.9495 & 0.9501 & 0.9253 & $+$0.5194 \\
DDoS\_HTTP            & 0.4027 & 0.7607 & 0.5843 & 0.6700 & $+$0.2673 \\
Ransomware            & 0.3862 & 0.6644 & 0.2636 & 0.6370 & $+$0.2508 \\
Uploading             & 0.1474 & 0.8439 & 0.4592 & 0.3738 & $+$0.2264 \\
Fingerprinting        & 0.1000 & 0.1000 & 0.1000 & 0.2267 & $+$0.1267 \\
Password              & 0.0723 & 0.0444 & 0.0146 & 0.1285 & $+$0.0562 \\
\midrule
DDoS\_UDP             & 1.0000 & 0.9693 & 0.9926 & 1.0000 & $+$0.0000 \\
MITM                  & 0.3132 & 0.2308 & 0.3132 & 0.3132 & $+$0.0000 \\
Normal                & 1.0000 & 0.9993 & 1.0000 & 0.9999 & $-$0.0001 \\
DDoS\_ICMP            & 0.9998 & 0.9940 & 0.9993 & 0.9995 & $-$0.0003 \\
SQL\_injection        & 0.9060 & 0.1025 & 0.8405 & 0.9017 & $-$0.0043 \\
XSS                   & 0.0637 & 0.1290 & 0.0000 & 0.0570 & $-$0.0067 \\
Vulnerability\_scanner& 0.7623 & 0.7355 & 0.8487 & 0.7435 & $-$0.0188 \\
DDoS\_TCP             & 0.9533 & 0.1557 & 0.8382 & 0.9226 & $-$0.0306 \\
\midrule
Backdoor              & 0.8965 & 0.0000 & 0.9150 & 0.7868 & $-$0.1097 \\
\bottomrule
\end{tabular}

\smallskip
\raggedright\footnotesize
$\Delta$ shows change VARS-FL vs.\ FedAvg.
Rows grouped: top (above first rule) = VARS-FL gain $> +0.05$,
middle = $|\Delta| \leq 0.05$,
bottom = VARS-FL loss $< -0.05$.
\end{table}

\textcolor{black}{\subsection{Robustness to Validation Set Composition}}
\label{sec:ablation-valset}
\textcolor{black}{
Given that VARS-FL relies on a server-side validation set to compute client scores, we investigate whether the validation-loss improvement signal $\delta_i^t$ in Eq.~\ref{eq:delta} depends on the \emph{composition} of the server-side validation set $\mathcal{D}_{val}$. Since practical deployments usually have access to a small balanced validation set with a fixed number of attack samples per class collected in a controlled setting, rather than large stratified holdouts, we evaluate VARS-FL under a uniform validation set(equal samples per class) to test the robustness of the selection mechanism.}

\begin{figure}[t]
    \centering
    \includegraphics[width=\linewidth]{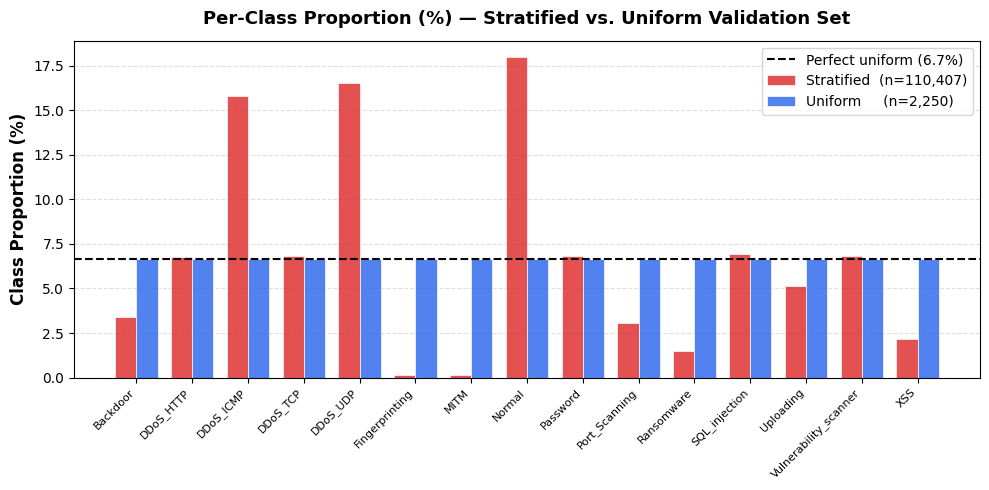}
    \caption{\textcolor{black}{
    Class data proportions for the stratified ($n=110{,}407$)
    and uniform ($n=2{,}250$) validation sets. The stratified set reflects the natural class imbalance of Edge-IIoTset. The uniform set assigns exactly $150$ samples per class and aligns with the perfect-uniform baseline (dashed line at $6.7\%$).}}
    \label{fig:val_ablation}
\end{figure}

\textcolor{black}{
In this ablation, we modify only the server-side validation set and run VARS-FL on the same three random seeds used in Section~\ref{sec:results}, keeping the test set fixed for both conditions to ensure a fair comparison of downstream performance. The original validation set used in our main experiments is \emph{stratified}, containing $110{,}407$ samples drawn proportionally to the class frequencies of Edge-IIoTset. The \emph{uniform} validation set is considerably smaller at $n=2{,}250$ samples ($150$ per class for all $15$ classes), as the per-class count is bounded by the rarest attack (\textit{Fingerprinting}). All other components of the training pipeline --training data, client partitions, model architecture, and hyperparameters---remain unchanged. The contrast in class distribution between the two validation sets is illustrated in Figure~\ref{fig:val_ablation}.}

\textcolor{black}{
The results of the validation set comparison are reported in Table~\ref{tab:val-ablation}. The table shows the test-set performance (mean~$\pm$~std over three seeds) under both validation set distributions. Between both configurations, all metric results shift by less than one percentage point. The accuracy changes by only $0.24$~pp, showing that the overall performance of the model is not significantly affected by the change in validation set composition. Most notably, F1-Macro, being the most likely to degrade if $\delta_i^t$ were biased toward majority classes, changes by only $0.48$~pp, well within the seed-to-seed standard deviation of $\pm 2.41$~pp observed in our main results. The validation loss shows barely any change ($\Delta = -0.0049$), and variance decreases with the uniform distribution (std drops from $0.0200$ to $0.0136$). This is consistent with reduced class imbalance noise on the smaller balanced distribution.}

\begin{table}[H]
\centering
\caption{VARS-FL Test Performance Under Stratified vs.\ Uniform
Validation Sets Used for Client Scoring (3 Seeds, 100 Clients,
100 Rounds). The Test Set Is Identical Across Both Columns.}
\label{tab:val-ablation}
\setlength{\tabcolsep}{4pt}
\renewcommand{\arraystretch}{1.1}
\small
\begin{tabular}{lccc}
\toprule
\textbf{Metric} & \textbf{Stratified Val} & \textbf{Uniform Val} & \textbf{$\Delta$} \\
                & ($n=110{,}407$)         & ($n=2{,}250$)        &                   \\
\midrule
Accuracy            & $\mathbf{0.8185 \pm 0.0040}$ & $0.8161 \pm 0.0071$ & $-0.0024$ \\
F1-Macro            & $\mathbf{0.6422 \pm 0.0241}$ & $0.6374 \pm 0.0278$ & $-0.0048$ \\
F1-Weighted         & $\mathbf{0.8086 \pm 0.0080}$ & $0.8041 \pm 0.0110$ & $-0.0045$ \\
Precision           & $\mathbf{0.8026 \pm 0.0392}$ & $0.7391 \pm 0.0466$ & $-0.0635$ \\
Loss ($\downarrow$) & $0.4937 \pm 0.0200$          & $\mathbf{0.4888 \pm 0.0136}$ & $-0.0049$ \\
\bottomrule
\end{tabular}
\end{table}

\textcolor{black}{
The key to this robustness is that the scoring mechanism ranks clients by their \emph{relative} loss improvement for each client and not their absolute loss values. The formulation of $\delta_i^t$ itself describes the score as a \emph{difference} of two losses computed on the \emph{same} validation set (the loss before the client's update and the loss after). Therefore, any distributional bias present in $\mathcal{L}_{\text{val}}(\theta^{t-1})$ also appears in $\mathcal{L}_{\text{val}}(\theta_i^t)$ and largely cancels. Only the \emph{relative} loss improvement of each client remains, which is the signal needed to rank clients. This shows that class distribution of validation set has minimal impact on VARS-FL scoring, as long as it is large enough that loss changes can be reliably detected above sampling noise.}

\textcolor{black}{
This finding is significant for deployment because it shows that 
operators do not need to maintain a large stratified validation set 
that tracks the (often unknown or non-stationary) global attack 
distribution. This is well-suited for IIoT environments, where the 
threat landscape evolves over time. Furthermore, the server validation 
set must be constructed independently from a limited trusted source 
rather than aggregated from client data, respecting federated 
learning's core privacy constraint where clients cannot expose raw 
data even to the server. These results confirm that VARS-FL is robust 
to the choice of validation set, as a small class-balanced set of a 
few thousand samples, collected independently from client devices, is 
sufficient to drive the selection mechanism reliably. Overall, this 
strengthens the practical feasibility of VARS-FL for IIoT deployments, 
where the true class distribution is hard to estimate and data 
sovereignty requirements prevent the server from accessing client 
traffic directly}

\section{Discussion and Conclusion}
\label{sec:conclusion}

We presented VARS-FL, a client selection framework that addresses the limitations of stateless and proxy-based strategies through a validation-aligned, accumulation-based Reputation mechanism. By combining a sliding-window average of recent validation-loss improvements with a logarithmically scaled participation term, VARS-FL captures both short-term utility and long-term reliability, enabling an effective balance between exploration and exploitation while preventing the exclusion of infrequent but informative clients. Defining the selection reward as global validation-loss reduction ensures alignment with the optimization objective, avoiding the misalignment issues inherent to local proxy metrics under non-IID data. Experimental results on a 15-class intrusion detection task show that VARS-FL consistently improves convergence speed, stability, and overall performance, reducing the rounds required to reach target accuracy by up to 36\% and achieving lower test loss with minimal cross-seed variance compared to FedAvg, Oort, and Power-of-Choice. A key limitation is the reliance on a server-side validation set; while results demonstrate robustness to class imbalance, sufficient coverage of all classes remains necessary, as unseen classes may be undervalued. Future work will explore coverage-aware scoring, theoretical convergence guarantees, and adaptive reputation mechanisms. Overall, VARS-FL reframes client selection as an objective-aligned, evidence-driven process, providing a principled and practical solution for federated learning under heterogeneous and non-IID data distributions.


\section*{Acknowledgment}
This work was supported by the United Arab Emirates University under Grant (G00005769).

\bibliographystyle{IEEEtran}
\bibliography{references}

\end{document}